\documentclass[10pt,twocolumn,letterpaper]{article}
\usepackage{graphicx}
\usepackage{wacv}      

\usepackage{graphicx}
\usepackage{amsmath}
\usepackage{amssymb}
\usepackage{booktabs}

\usepackage[pagebackref,breaklinks,colorlinks]{hyperref}

\usepackage[capitalize]{cleveref}
\crefname{section}{Sec.}{Secs.}
\Crefname{section}{Section}{Sections}
\Crefname{table}{Table}{Tables}
\crefname{table}{Tab.}{Tabs.}

\def\wacvPaperID{1453} 
\def\confName{WACV}
\def\confYear{2024}

\begin{document}

\title{IMPROVE: Improving Medical Plausibility without Reliance on Human Validation - An Enhanced Prototype-Guided Diffusion Framework}
\author{
Anurag Shandilya \and
Swapnil Bhat \and
Akshat Gautam \and
Subhash Yadav \and
Siddharth Bhatt \and
Deval Mehta \and
Kshitij Jhadav \\
}
\maketitle

\begin{abstract}
Generative models have proven to be very effective in generating synthetic medical images and find applications in downstream tasks such as enhancing rare disease datasets, long-tailed dataset augmentation, and scaling machine learning algorithms. For medical applications, the synthetically generated medical images by such models are still reasonable in quality when evaluated based on traditional metrics such as FID score, precision, and recall. However, these metrics fail to capture the medical/biological plausibility of the generated images. Human expert feedback has been used to get biological plausibility which demonstrates that these generated images have very low plausibility. Recently, the research community has further integrated this human feedback through Reinforcement Learning from Human Feedback(RLHF), which generates more medically plausible images. However, incorporating human feedback is a costly and slow process. In this work, we propose a novel approach to improve the medical plausibility of generated images without the need for human feedback. We introduce \textbf{IMPROVE}: \textbf{I}mproving \textbf{M}edical \textbf{P}lausibility without \textbf{R}eliance \textbf{o}n Human \textbf{V}alidation - An \textbf{E}nhanced Prototype-Guided Diffusion Framework, a prototype-guided diffusion process for medical image generation and show that it substantially enhances the biological plausibility of the generated medical images without the need for any human feedback. We perform experiments on Bone Marrow and HAM10000 datasets and show that medical accuracy can be substantially increased without human feedback.

\end{abstract}

\section{Introduction}
\label{sec:intro}

The absence of large-scale annotated images proves to be a big hurdle in the applicability of deep learning algorithms in medicine\cite{chen2021synthetic}. The difficulty arises from regulatory hurdles that prevent data sharing, the cost involved in getting medical experts’ feedback which can be prohibitive\cite{budd2021survey}, and the fact that many medical conditions are rare in occurrence\cite{razzak2018deep} . Furthermore, the long-tailed nature of many medical datasets complicates the learning process for machine learning and deep learning algorithms\cite{wu2024medical} . Recently, Diffusion models have shown great promise in conditional/non-conditional generation of high-quality natural and artistic images\cite{packhauser2023generation, ali2022spot} .  These synthetically generated images can potentially alleviate the issues by augmenting medical datasets and balancing long-tailed datasets.

Traditional image generation pipelines prioritize the visual or artistic appeal of the generated images. Unlike these cases where visual realism is of prime focus, medical image generation requires grounding in fundamental biological and clinical realities. Therefore, there is very little room for an unrestricted or creative generation. Downstream application of these images in critical operation demands an accurate description of the anatomy, pathology as well as other biological constraints.  Generative models traditionally have no context of medical/clinical concepts and are therefore inept at generating medically/biologically plausible images\cite{adams2023does, chambon2022adapting}.  Thus whenever they are used for medical image generation, a high fraction of biologically implausible images are generated\cite{sun2024aligning}. This entails that more expert manpower needs to be deployed in order to obtain the required collection of medically plausible images. Therefore, our aim is to develop a framework for generating medical images that exceeds traditional approaches in medical plausibility.

Foundation models such as Stable Diffusion have been trained on a very large corpus of natural images (LAION\_5B dataset)\cite{schuhmann2022laion} . One approach for synthetic medical image generation could be to fine-tune these models on real medical images\cite{ali2022spot} . With sufficient medical images, we can expect a fine-tuned model to capture the latent biological knowledge from medical images. However, most of the medical datasets are relatively smaller(thousands in size), and getting experts’ annotations is costly. If we happen to train a large vision model on such a dataset with a classical objective function, the model fails to capture biological information. Designing domain-specific objective functions for the fine-tuning task is challenging and does not generalize\cite{sun2024aligning}. On the other hand, a conditional diffusion model conditioned on classes or class description fails to capture the true distribution of biological information\cite{sun2024aligning} . 

Recently, there has been an attempt to integrate user feedback through RLHF in order to increase the medical plausibility of generated images\cite{sun2024aligning} . However, that still leaves us with a two-stage process with the second stage involving experts' feedback which is costly and time-consuming. 

In this work, we demonstrate that the medical plausibility of generated images can be increased substantially without the need for experts' feedback. This could dramatically reduce the cost and time required for medical image generation.

\textbf{Summary of our key contributions:} \\
1) We propose a new prototype guided diffusion model-based medical image generation pipeline that increases the biological accuracy of medical images generated.\\
2) We generate images for bone-marrow and dermatology datasets and demonstrate that increase in biological plausibility is consistent across two different domains.


\section{Related Works}
\label{sec:formatting}
\subsection{Prototype Learning}
Recently, prototype learning utilizing deep networks has caught the attention of researchers and has been applied for a range of tasks through enhancing of the feature space by learning clusters (prototypes) of different classes/categories. Zhou et al \cite{zhou2022rethinking} improved upon semantic segmentation through the representation of different semantic classes with prototypes. Snell et al \cite{snell2017prototypical} proposed prototype learning for a few-shot classification problem whereas Shu et al \cite{shu2020p} devised a prototypical network for an open set recognition task. Prototypes have been integrated into both supervised \cite{wu2018unsupervised} as well as unsupervised \cite{guerriero2018deepncm} classification tasks. Kim et al \cite{kim2019variational} learn a discrete space that included prototypes for image reconstruction tasks. Improvements have been made in prototype learning architecture as well. Du et al \cite{du2023protodiff} introduce ProtoDiff, a novel framework that leverages a task-guided diffusion model during the meta-training phase to gradually generate prototypes, thereby providing efficient class representations.

\subsection{Diffusion Models}
Diffusion models are generative models that have recently shown great promise in their capability to generate high-fidelity images for a variety of applications. Decreased sampling times \cite{song2020denoising, salimans2022progressive} and architectural improvements such as cascaded diffusion \cite{ho2022cascaded} have been proposed to improve the generative ability of diffusion models. In parallel, a new area of research has focused on conditioning the diffusion models through text and images. This conditioning helps in maintaining control over the diffusion process and results in the improvement of generation quality. Traditionally, the conditioning information is embedded in the latent space through an encoder such as CLIP\cite{radford2021learning} or VQGAN \cite{esser2021taming}. Retrieval Augmented Diffusion\cite{blattmann2022retrieval} conditions the generation process on the neighbors of an image.  Our method has a similar intuition as learned prototypes have a similar advantage in capturing latent information of a class as neighboring images may have.

\subsection{Diffusion Models for medical image generation}
The traditional approach for medical image generation has been to finetune a diffusion model using sufficient samples from a given modality. Diffusion probabilistic models\cite{dhariwal2021diffusion} have recently shown great promise in the generation of high-fidelity medical images\cite{muller2023multimodal}. Ali et al \cite{ali2022spot} showed that Stable Diffusion can generate high-quality X-Ray and CT images. Latent diffusion models \cite{pinaya2022brain} have been used for generating synthetic images from 3D brain images. 

Traditionally, these generative models for medical images are evaluated either indirectly through downstream classification tasks or directly through metrics such as FID scores\cite{heusel2017gans}. Other approaches involve sample-level evaluation metrics such as precision and recall\cite{sajjadi2018assessing} which check if synthetic samples reside in support of real data distribution. However, these approaches fail to incorporate biological/medical information\cite{sun2024aligning} which is of paramount importance while evaluating the medical plausibility of generated images. There are some traditional metrics for medical image evaluation such as signal-to-noise ratio and contrast-to-noise ratio \cite{winkler2008evolution} which are applied to real images and cannot be repurposed for evaluation of synthetic medical images. Evaluating synthetic images could be quite challenging as we do not have any ground truth as to what is medically plausible and what is not. Therefore it becomes imperative to design robust evaluation criteria that are domain-specific and capture different dimensions specific to the domain. In the medical domain, this entails getting expert feedback to ascertain the biological/medical plausibility of generated images. Sun et al \cite{sun2024aligning} were the first to introduce human feedback as the gold standard for the evaluation of synthetically generated images. Further, they showed that the biological plausibility of the generated images increases dramatically once human feedbacks are integrated through RLHF.


\section{Background}
\subsection{Diffusion Models}
Gaussian diffusion models have become a prominent generative method since their introduction by \cite{sohl2015deep}. These models have inspired numerous variants based on the diffusion concept. During the training phase, Gaussian noise is incrementally added to a sample from the real data distribution $x_0 \sim q(x)$ over T time steps:
\begin{equation} \label{eq:forward_diffusion_beta}
q(x_t \mid x_{t-1}) = \mathcal{N}(x_t; \sqrt{1 - \beta_t} x_{t-1}, \beta_t I)
\end{equation}
Here, $x_{t-1}$ is transformed into $x_t$ by adding Gaussian noise at the 
$t-th$ time step. The variance of the added noise is $\beta_t$ and $\sqrt{1-\beta_t}$ is the scaling parameter based on a variance schedule. To obtain $x_t$ at any arbitrary time step t without iterating t steps, a reparameterization trick, leveraging properties of the Gaussian distribution, can be applied:
\begin{equation} \label{eq:forward_diffusion_alpha}
q(x_t \mid x_0) = \mathcal{N}(x_t; \sqrt{\bar{\alpha}_t} x_0, (1 - \bar{\alpha}_t) I)
\end{equation}
\begin{equation} \label{eq:reparametarization_trick}
x_t = \sqrt{\bar{\alpha}_t} x_0 + \sqrt{1 - \bar{\alpha}_t} \epsilon
\end{equation}
In these equations, $\alpha_t = 1 - \beta_t, \quad \bar{\alpha}_t = \prod_{i=1}^{t} \alpha_i, \quad \text{and} \quad \epsilon \sim \mathcal{N}(0, I)$
The training phase of diffusion models involves a denoising process to reverse the noising process in Equation \ref{eq:forward_diffusion_beta}. This denoising process is defined as:
\begin{equation}\label{eq:reverse_diffusion_process}
p_\theta(x_{t-1} \mid x_t) = \mathcal{N}(x_{t-1}; \mu_\theta(x_t, t), \Sigma_\theta(x_t, t))
\end{equation}
Here, $\theta$ represents the parameters of a neural network that predicts the mean $\mu_{\theta}(x_t,t)$ and variance $\Sigma_{\theta}(x_t,t)$ of the Gaussian distribution. Starting from pure Gaussian noise $x_T$, the image $x_0$ is obtained by progressively reducing the noise over T time steps. The main goal of diffusion models as generative models is to learn this reverse process to generate a high-quality image $x_0$ from random noise $x_T$. \\
In Denoising Diffusion Probablistic Models(DDPM) \cite{ho2020denoising}, the mean $\mu_{\theta}(x_t,t)$ is learned while the variance $\Sigma_{\theta}(x_t,t)$ is kept constant. A tractable variational lower bound exists for optimizing the neural network in Equation \ref{eq:reverse_diffusion_process}.\cite{ho2020denoising} decompose the objective function and demonstrate that predicting the noise $\epsilon$ added in the current time step, as in Equation \ref{eq:forward_diffusion_alpha}, is the optimal way to parameterize the model’s mean $\mu_{\theta}(x_t,t)$: 
\medskip

\begin{equation}
\mu_\theta(x_t, t) = \frac{1}{\sqrt{\alpha_t}} \left( x_t - \frac{\beta_t}{\sqrt{1 - \bar{\alpha}_t}} \epsilon_\theta(x_t, t) \right)
\end{equation}

\medskip

The simplified training objective is derived as:

\medskip

\begin{equation}
L_\text{simple} = \mathbb{E}_{t \sim [1,T], x_0 \sim q(x), \epsilon \sim \mathcal{N}(0, I)} \left[ \| \epsilon - \epsilon_\theta(x_t, t) \|^2 \right]
\end{equation}

\medskip

While DDPM generates unconditional images, guided diffusion models are also used for conditional image generation. \cite{dhariwal2021diffusion} propose classifier guidance, where the class-conditional parameters $\mu_{\theta}(x_t \mid y)$ and $\Sigma_{\theta}(x_t \mid y)$ are adjusted using the gradients of a classifier $p_{\phi}(y\mid x_t)$ predicting the target class $y$.  The perturbed mean with the guidance scale $s$ is:

\medskip

\begin{equation}
\mu_\theta(x_t \mid y) = \mu_\theta(x_t \mid y) + s \Sigma_\theta(x_t \mid y) \nabla_{x_t} \log p_\phi(y \mid x_t)
\end{equation}

\medskip

Despite improving image quality, classifier guidance has its challenges. As the denoising process starts with highly noisy input and proceeds with noisy images for most time steps, the classifier must be robust to noise. Obtaining such a classifier is difficult, and predicting a class label does not require most of the data's information, potentially misguiding the generation direction.

\cite{ho2022classifier} propose a classifier-free guidance method, eliminating the need for a separate classifier. The conditioning information $y$ is periodically used, and dropped out at other times, allowing a single model for both unconditional and conditional generation. They derive that unconditional $\epsilon_{\theta}(x_t,t)$ and conditional $\epsilon_{\theta}(x_t,t,y)$ estimations represent the classifier's gradients as:

\begin{multline}\label{eq:classifier_free_guidance}
\nabla_{x_t} \log p(y \mid x_t) = \nabla_{x_t} \log p(x_t \mid y) - \nabla_{x_t} \log p(x_t) \\
= -\frac{1}{\sqrt{1 - \bar{\alpha}_t}} (\epsilon_\theta(x_t, t, y) - \epsilon_\theta(x_t, t))
\end{multline}

Equation \ref{eq:classifier_free_guidance} suggests that an implicit classifier can replace the need for an explicit one, with \cite{ho2022classifier} reporting better results with classifier-free guidance.\\
Our work utilizes the method proposed by \cite{baykal2024protodiffusion} to improve the generation performance by utilizing prototype learning.
\subsection{Prototype Learning}

Prototype learning is a foundational concept in machine learning that enhances the interpretability and robustness of classification models by utilizing representative examples, or prototypes, for each class. These prototypes serve as central reference points that capture the essential characteristics of their respective classes. Unlike traditional methods that rely solely on raw data or hand-engineered features, prototype learning integrates these learned prototypes into the classification process, thereby improving both accuracy and resilience to unseen data.

The study draws upon the pioneering work of~\cite{yang2018robust}, which introduced convolutional prototype learning (CPL) within deep convolutional neural networks (CNNs). Yang et al. proposed a framework where CNNs act as feature extractors \( f(x;\theta) \), extracting high-level features from raw medical images \( x \). Rather than employing a softmax layer for classification, CPL maintains and learns several prototypes \( m_{ij} \) for each class \( i \), enhancing the model’s ability to generalize beyond the training data.

In the CPL framework, the distance is used to measure the similarity between the samples and the prototypes. Consequently, the probability of a sample \((x, y)\) belonging to the prototype \( m_{ij} \) is determined by the distance between them:
\begin{equation}
p(x \in m_{ij} | x) \propto - \| f(x) - m_{ij} \|^2_2.
\end{equation}
To ensure the probability is non-negative and sums to one, the probability \( p(x \in m_{ij} | x) \) is defined as:
\begin{equation}
p(x \in m_{ij} | x) = \frac{e^{-\gamma d(f(x), m_{ij})}}{\sum_{k=1}^C \sum_{l=1}^K e^{-\gamma d(f(x), m_{kl})}},
\end{equation}
where \( d(f(x), m_{ij}) = \| f(x) - m_{ij} \|^2_2 \) represents the distance between \( f(x) \) and \( m_{ij} \). \( \gamma \) is a hyper-parameter that controls the hardness of probability assignment. Given this definition, the probability \( p(y | x) \) is expressed as:
\begin{equation}
p(y | x) = \sum_{j=1}^K p(x \in m_{yj} | x).
\end{equation}
Based on the probability \( p(y | x) \), the cross entropy (CE) loss in the CPL framework is defined as:
\begin{equation}
l((x, y); \theta, M) = -\log p(y | x).
\end{equation}
This loss function is based on distance, differentiating it from traditional cross-entropy loss, and is thus referred to as distance-based cross-entropy (DCE) loss. From Equations (12), (13), and (14), minimizing the loss function essentially reduces the distance between the samples and the prototypes of their true classes. The DCE is also derivable with respect to \( M \) and \( f \).

Additionally, a prototype loss (PL) is introduced to refine the optimization process by decreasing distances between features and their corresponding prototypes, thereby promoting both intra-class compactness and inter-class separability. The prototype loss is defined as:
\begin{equation}
pl((x, y); \theta, M) = \| f(x) - m_{yj} \|^2_2,
\end{equation}
where \( m_{yj} \) is the closest prototype to \( f(x) \) from the corresponding class \( y \). The prototype loss is combined with the classification loss to train the model, resulting in the total loss:
\begin{equation}
\text{loss}((x, y); \theta, M) = l((x, y); \theta, M) + \lambda pl((x, y); \theta, M),
\end{equation}
where \( \lambda \) is a hyper-parameter controlling the weight of the prototype loss.

By integrating these losses, the CPL framework achieves intra-class compactness and inter-class separability, which are crucial for robust classification and generalization to new classes.

\begin{figure}[h]
    \centering
    \includegraphics[width=0.5\textwidth]{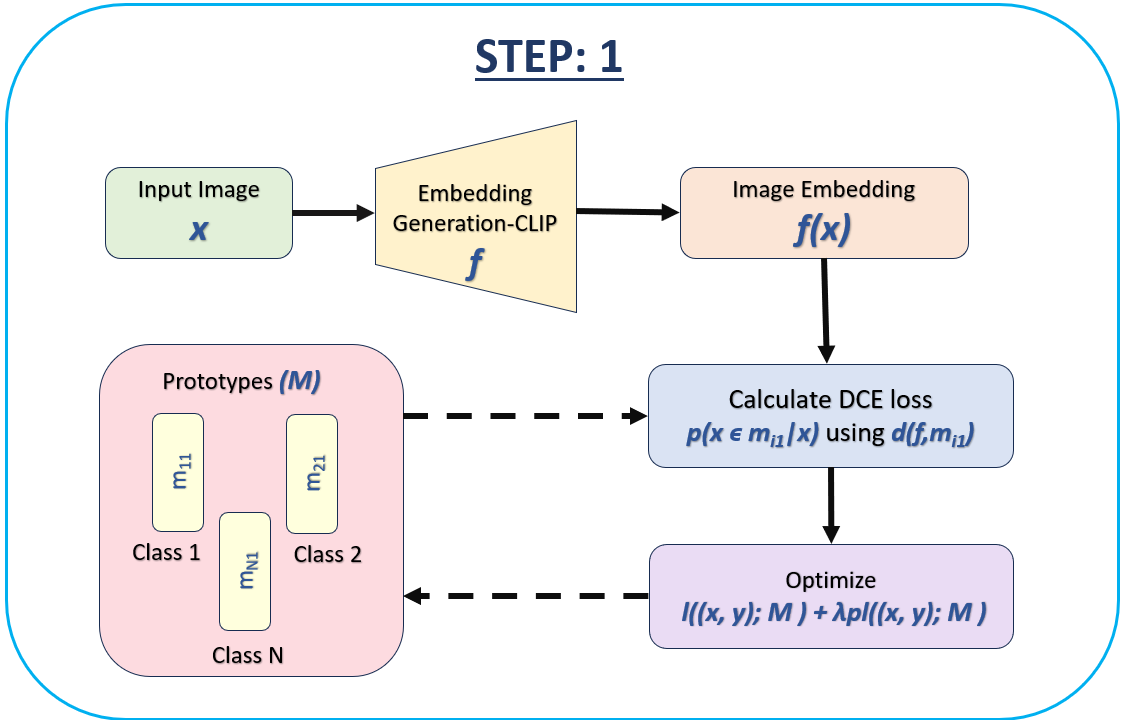} 
    \caption{Prototype generation represents the first step of the training process, here a Codebook(\textit{M}) of prototypes is learned}
    \label{fig:proto_model_step_1}
\end{figure}

\section{Proposed Method}
Our proposed method leverages prototype-guided diffusion models for zero-shot conditional image generation. We use predefined prototypes to guide the diffusion process, ensuring that the generated images adhere closely to medically relevant characteristics. These prototypes act as high-level representations of the target image features and are supplied as inputs to the diffusion model.
We learn the prototypes of the classes with a different classifier first, and then we start the training of the diffusion model after we initialize the class embeddings with the learned prototypes which are used to guide the diffusion process.\\
Our model achieves better performance than the original diffusion model when the quality of generated images is compared in terms of clinical validity and biological accuracy. 
\subsection{Prototype Generation}
Each Dataset contains $N$ classes- from each class an equal number of images $n$ are selected for training.\\
The first step involves extracting the image embeddings using a pretrained CLIP model. The dimension of an embedding is 768. These image embeddings are utilized in the training process as shown in Figure \ref{fig:proto_model_step_1} where the prototype representations of each class are learned. The learned prototypes have the same dimension as the input image embeddings. These prototypes differ from the image embeddings themselves in the sense that they provide a representation of the entire class demonstrating intra-class compactness as well as enhancing the differences in feature representation between different classes- inter-class separability. We claim that the prototypes capture the biological intricacies of the different structures in a much better, and using these prototypes for guiding the diffusion process would result in better results than classifier-free guidance using only class labels.

\subsection{Diffusion Model Architecture}
\begin{figure*}[t]
    \centering
    \includegraphics[height=0.3\textheight]{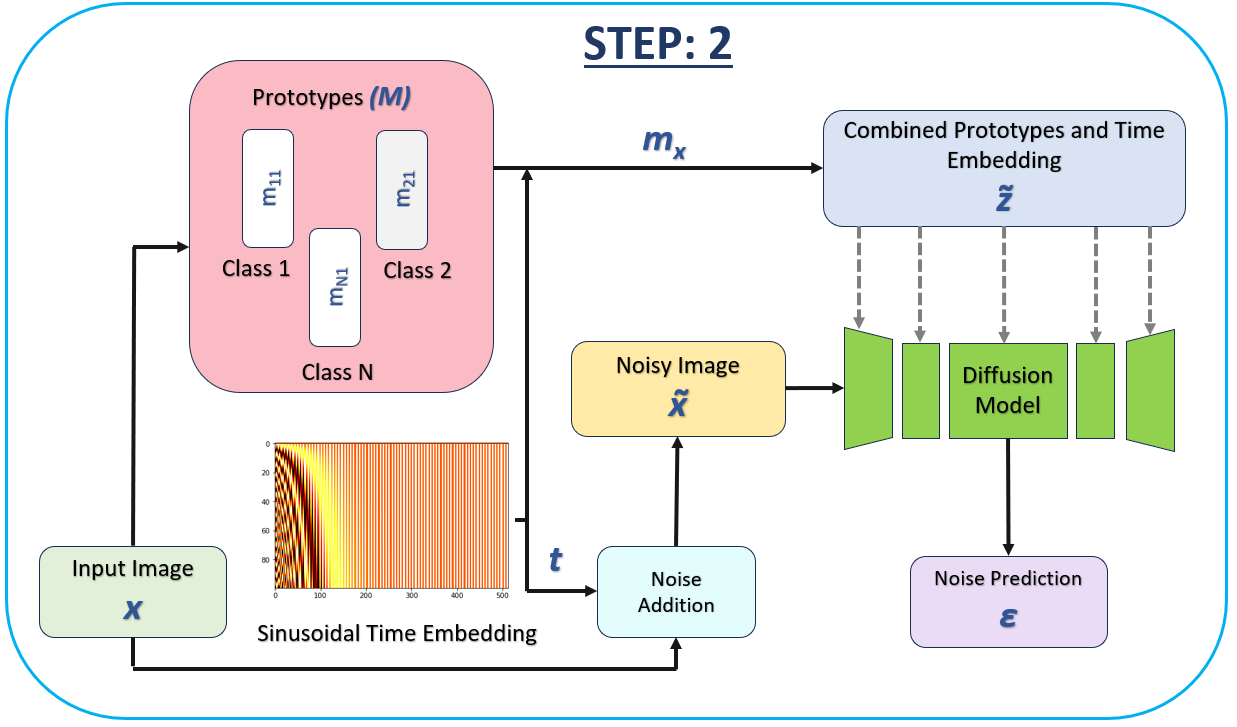} 
    \caption{Time embeddings and learnt prototypes are concatenated and passed at various stages in the UNET model to guide the diffusion process}
    \label{fig:proto_model_step_2}
\end{figure*}

Our diffusion model architecture follows closely the method proposed by \cite{baykal2024protodiffusion}, where we initialize the class embeddings with the learned prototypes as shown in Figure \ref{fig:proto_model_step_2}. The difference is that we freeze the weights of the class embeddings and they don't change during the training process. We are concatenating the sinusoidal time embedding t with the prototype $m_x$ that represents the class label of the input $x$, and obtain $\tilde{z}$. The core architecture comprises of a UNet-like neural network which is optimized to predict the noise $\epsilon$ of the noisy image $\tilde{x}$ added at each timestep $t$. Over a gradual denoising process, the network is able to produce an image from the learned probability distribution. In order to guide the diffusion process, the class embedding and the time step are also given to the specific layers of the neural network.
For the same model training parameters and hyperparameters, the baseline model has randomly initialized class embeddings which are optimized by the diffusion process during training without any explicit optimizations to improve their class representativeness. On the other hand, the prototype-guided model starts with optimized embeddings resulting in better results in a single shot.

\section{Experiments}
We conduct our experiments on the Bone marrow cell classification dataset and HAM10000 ("Human Against Machine with 10000 training images") dataset - a large collection of multi-source dermatoscopic images of pigmented lesions using 2 NVIDIA A100 GPUs.
\subsection{Bone Marrow dataset}
The Bone Marrow dataset~\cite{matek2021expert} comprises over 170,000 de-identified, expert-annotated cells derived from the bone marrow smears of 945 patients. These smears were stained using the May-Grünwald-Giemsa/Pappenheim stain. For our experiments, we utilize images categorized into 16 distinct classes, as detailed in Table \ref{tab:bone_marrow_classes}.
\begin{table}[h]
    \centering
    \caption{Classes in the Bone Marrow dataset}
    \label{tab:bone_marrow_classes}
    \begin{tabular}{|c|l|}
        \hline
        Abbreviation & Class Name \\
        \hline
        ART & Artefact \\
        BAS & Basophil \\
        BLA & Blast \\
        EBO & Erythroblast \\
        EOS & Eosinophil \\
        HAC & Hairy cell \\
        LYT & Lymphocyte \\
        MMZ & Metamyelocyte \\
        MON & Monocyte \\
        MYB & Myelocyte \\
        NGB & Band neutrophil \\
        NGS & Segmented neutrophil \\
        NIF & Not identifiable \\
        PEB & Proerythroblast \\
        PLM & Plasma cell \\
        PMO & Promyelocyte \\
        \hline
    \end{tabular}
\end{table}

\subsection{Skin Cancer MNIST: HAM10000}
The HAM10000 ("Human Against Machine with 10000 training images")~\cite{DVN/DBW86T_2018} dataset comprises a diverse collection of dermatoscopic images from various populations, acquired using different modalities. The images are categorized into several classes: actinic keratoses and intraepithelial carcinoma/Bowen's disease (akiec), basal cell carcinoma (bcc), benign keratosis-like lesions (solar lentigines, seborrheic keratoses, and lichen-planus like keratoses, bkl), dermatofibroma (df), melanoma (mel), melanocytic nevi (nv), and vascular lesions (angiomas, angiokeratomas, pyogenic granulomas, and hemorrhage, vasc).
\subsection{Image Generation and Feedback}
\begin{figure}[h]
    \centering
    \includegraphics[width=0.4\textwidth]{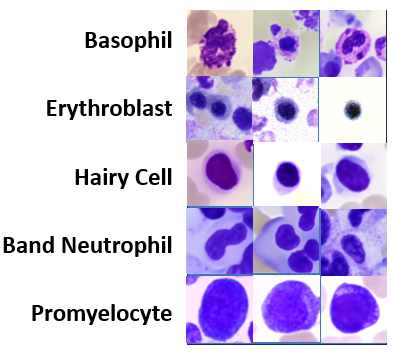} 
    \caption{Generated images for 5 classes of Bone Marrow Dataset using Classifier Free Guidance}
    \label{fig:baseline_bone_marrow}
\end{figure}

\begin{figure}[h]
    \centering
    \includegraphics[width=0.33\textwidth]{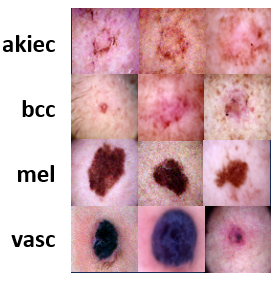} 
    \caption{Generated images for 4 classes of Skin Cancer Dataset using Prototype Guidance}
    \label{fig:prototype_skin_cancer}
\end{figure}

We used a prototype-guided diffusion model trained on real images (64×64 pixels) to generate synthetic image patches. For both datasets, Model training was conducted using 100 images per class. The prototypes are learned using the same images and then these learned prototypes are concatenated with the time embedding and fed as input to different layers of the UNET model. \\
For the baseline model, similar image size (64x64 pixels) were used to generate the synthetic image patches with 100 images per class. The class embeddings for the baseline are randomly initialized.\\
In both cases the models are initialized with same set of hyperparameters, values of some of the hyperparameters are mentioned in table \ref{tab:hyperparameters}.\\
The initial channel size of the UNET model is set to 64 and 4 blocks are utilized for the down and upsampling layers of the model. Each of them has 2 ResNet layers. These layers have a 0.1 Dropout layer. The number of channels keeps on increasing during the Encoder part of the model which is represented by the downsampling layers by a factor of 2,4 and 8. The Decoder part which contains the upsampling layers, mirrors the Encoder, and the channel size keeps on decreasing by the same factors. The final output of the UNET has the same number of channels as the input- 3. This type of hierarchical representation captures both high-level as well as fine features.\\
For training the diffusion model we employ AdamW optimizer with a learning rate of 0.0001. We resize the images to 64 for both datasets and a per GPU batch size of 16 is employed.\\
\begin{table}[h]
    \centering
    \caption{Hyperparameters of Diffusion model}
    \label{tab:hyperparameters}
    \begin{tabular}{|c|l|}
       \hline
        Name & Value \\
        \hline
        Input image size & 64 \\
        Initial channel size & 64 \\
        Channel multiplier & [1, 2, 4, 8] \\
        Timesteps & 1000 \\
        Initial learning rate & 1e-4 \\
        Number of epochs & 1500 \\
        \hline
    \end{tabular}
\end{table}
The sampling process involves the generation of 100 images from each class which utilizes faster DDIM sampling \cite{song2020denoising} over 50 iterations.
The generated images are shown in Figures \ref{fig:baseline_bone_marrow} and \ref{fig:prototype_skin_cancer}.\\
\begin{figure}[h]
    \centering
    \includegraphics[width=0.4\textwidth]{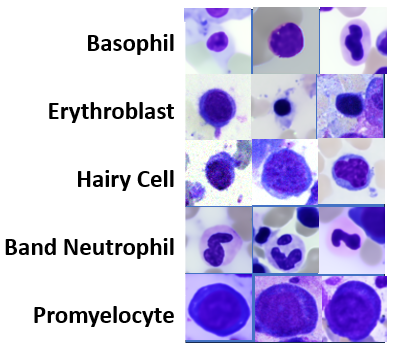} 
    \caption{Generated images for 5 classes of Bone Marrow Dataset using Prototype Guidance}
    \label{fig:prototype_bone_marrow}
\end{figure}
\begin{figure}[h]
    \centering
    \includegraphics[width=0.33\textwidth]{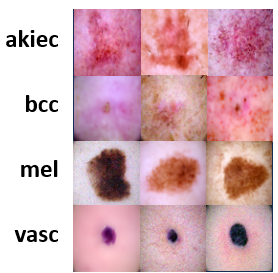} 
    \caption{Generated images for 4 classes of Skin Cancer Dataset using Classifier Free Guidance}
    \label{fig:baseline_skin_cancer}
\end{figure}
For feedback we relied on two domain experts; a pathologist for Bone Marrow dataset and a dermatologist for HAM10000 dataset. Labelstudio was used as a platform for gathering annotations. For Bone Marrow dataset, human expert was provided with plausibility criteria over which contribute to the biological viability of images. These criteria included cell size, nucleus shape \& size, nucleus-to-cytoplasm ratio, cytoplasm color and consistency, chromatin pattern, inclusions, and granules (where appropriate). Failure over any one or more criteria led to the categorization of the image as implausible. Similarly, for HAM10000, the dermatologist was provided with five criteria, all of which are needed to be satisfied for an image to be considered plausible.

\section{Results}

To evaluate the impact of our proposed architecture on the generation of medically plausible synthetic images, we created two synthetic datasets each for both Bone Marrow dataset and HAM10000 dataset, a sample from the classifier free guidance diffusion model(our baseline) and a sample from prototype guided diffusion model(our approach). For Bone Marrow, each of the two synthetic datasets had 1400 synthetic images(100 images per class). For HAM10000 the sampled quantity was 700 per dataset(here too, 100 images per class). Table~\ref{table:plausible_images_comparison_BM} lists the fraction of clinically plausible images per cell type for the two synthetic datasets for HAM10000 images as evaluated by an expert dermatologist. As we can see our model provides a significant bump in medical plausibility across all but one cell type. The average rate of clinical plausibility increased from 48.8 \% to 60 \%. \\

\begin{table}[h]
\centering
\caption{Comparison of plausibility percentage of synthetic images for HAM10000 dataset}
\label{table:plausible_images_comparison_BM}
\resizebox{\columnwidth}{!}{%
\begin{tabular}{|l|c|c|}
\hline
\textbf{Morphological Cell Type} & \textbf{Baseline (\%)} & \textbf{Our Approach (\%)} \\
\hline
Melanocytic Nevi & 68 & \textbf{79} \\
Melanoma & 48 & \textbf{65} \\
Benign K-like lesions & 46 & \textbf{57} \\
Basal Cell Carcinoma & 33 & \textbf{60} \\
Actinic Keratosis & 44 & \textbf{49} \\
Vascular Lesions & \textbf{54} & 47 \\
Dermatofibroma & 49 & \textbf{63} \\
\hline
\end{tabular}%
}
\end{table}

Table \ref{table:plausible_images_morphological_types} lists the fraction of clinically plausible images per cell type for the two synthetic datasets for Bone Marrow images. For all except two classes we see a substantial increase in the medical plausibility of medical images generated. Furthermore, certain cell types such as hairy cells, basophils, and plasma cells are extremely tricky given the nuanced features of each cell type. With these cells, the baseline model substantially struggles to produce medically plausible images; however, we observe a significant improvement through our approach. The average rate of clinical plausibility increased from 48.07 \% to 62.57 \%.
\begin{table}[h]
\centering
\caption{Comparison of plausibility percentage of synthetic images for Bone-Marrow dataset}
\label{table:plausible_images_morphological_types}
\resizebox{\columnwidth}{!}{%
\begin{tabular}{|l|c|c|}
\hline
\textbf{Morphological Cell Type} & \textbf{Baseline (\%)} & \textbf{Our Approach (\%)} \\
\hline
Plasma Cell & 34 & \textbf{56} \\
Erythroblast & 66 & \textbf{68} \\
Promyelocyte & 48 & \textbf{62} \\
Hairy Cell & 21 & \textbf{57} \\
Metamyelocyte & 39 & \textbf{53} \\
Band Neutrophil & 52 & \textbf{69} \\
Proerythroblast & \textbf{67} & 64 \\
Myelocyte & \textbf{57} & 56 \\
Monocyte & 44 & \textbf{64} \\
Lymphocyte & 58 & \textbf{63} \\
Eosinophil & 42 & \textbf{71} \\
Segmented Neutrophil & 74 & \textbf{78} \\
Blast Cell & 45 & \textbf{59} \\
Basophil & 26 & \textbf{56}  \\
\hline
\end{tabular}%
}
\end{table}

Accuracy in downstream tasks is one of the ways in evaluating the efficacy of synthetic image generation. We evaluated the utility of our model in training a cell-type classification model. In this experiment, we train a ResNext-50 model in classifying 14 cell types for Bone Marrow dataset and seven cell types for HAM1000 dataset once using the classifier free guidance dataset (baseline approach) and then again using our approach. To ensure a fair comparison, the size of synthetic datasets was kept the same. Thereafter, the classification accuracy of both models was tested on a held-out real dataset, containing 50 images per cell type. Classifier trained on real images are used as the baseline. Results are shown in table \ref{table:performance_comparison} and table \ref{table:performance_comparison1} for Bone Marrow and HAM10000 images respectively. As expected, the performance of classification over real images is the best across all metrics showing a significant gap compared to the classifier trained on synthetic images generated using classifier free guidance(baseline). However, our approach helped reduce this gap across both datasets to a point where the classifier trained on real images only marginally outperformed the ones trained on our synthetic images.

\begin{table}[h]
\centering
\caption{Performance of a classifier trained over real and synthetic images of Bone Marrow dataset}
\label{table:performance_comparison}
\resizebox{\columnwidth}{!}{%
\begin{tabular}{|l|c|c|c|}
\hline
\textbf{Method} & \textbf{Precision (\%)} & \textbf{Recall (\%)} & \textbf{F1 Score (\%)} \\
\hline
Baseline (Classifier Free Guidance) & 61.50 & 65.43 & 63.40 \\
Ours & 72.65 & 76.64 & 74.58 \\
Real Images & 80.11 & 79.23 & 79.66 \\
\hline
\end{tabular}%
}
\end{table}

\begin{table}[h]
\centering
\caption{Performance of a classifier trained over real and synthetic images of HAM10000 dataset}
\label{table:performance_comparison1}
\resizebox{\columnwidth}{!}{%
\begin{tabular}{|l|c|c|c|}
\hline
\textbf{Method} & \textbf{Precision (\%)} & \textbf{Recall (\%)} & \textbf{F1 Score (\%)} \\
\hline
Baseline (Classifier Free Guidance) & 68.95 & 73.43 & 71.10 \\
Ours & 75.60 & 74.54 & 75.08 \\
Real Images & 84.56 & 87.32 & 85.91 \\
\hline
\end{tabular}%
}
\end{table}

\section{Conclusion}
Our prototype-guided diffusion framework represents a significant advancement in the field of synthetic medical image generation. By effectively eliminating the need for extensive human feedback, our approach not only reduces the associated costs and time but also produces synthetic images with high medical and biological accuracy. This work paves the way for more efficient and scalable solutions in medical data augmentation, ultimately contributing to the advancement of machine learning in healthcare.
Further, Our findings underscore the importance of integrating domain-specific knowledge into generative models and highlight the potential of prototype learning in enhancing the plausibility of synthetic medical images. We believe that our approach will inspire further innovations in the generation of high-quality synthetic medical data, thereby supporting the development of more robust and reliable AI-driven healthcare solutions.

{\small
\bibliographystyle{ieee_fullname}
\bibliography{egbib}
}

\end{document}